\documentclass[10pt, a4paper]{article}
\usepackage{lrec2022} 
\usepackage{multibib}
\newcites{languageresource}{Language Resources}
\usepackage{graphicx}
\usepackage{tabularx}
\usepackage{soul}
\usepackage{titlesec}
\titleformat{\section}{\normalfont\large\bfseries\center}{\thesection.}{1em}{}
\titleformat{\subsection}{\normalfont\SmallTitleFont\bfseries\raggedright}{\thesubsection.}{1em}{}
\titleformat{\subsubsection}{\normalfont\normalsize\bfseries\raggedright}{\thesubsubsection.}{1em}{}
\renewcommand\thesection{\arabic{section}}
\renewcommand\thesubsection{\thesection.\arabic{subsection}}
\renewcommand\thesubsubsection{\thesubsection.\arabic{subsubsection}}

\usepackage{epstopdf}
\usepackage[utf8]{inputenc}

\usepackage{hyperref}
\usepackage{xstring}
\usepackage{latexsym}
\usepackage{graphicx}
\usepackage{booktabs}
\usepackage{url}
\usepackage{microtype}
\usepackage{enumitem}
\usepackage{kotex}

\usepackage{color}

\title{StyleKQC: A Style-Variant Paraphrase Corpus for \\Korean Questions and Commands}

\name{Won Ik Cho$^1$, Sangwhan Moon$^2$$^,$$^3$, Jong In Kim$^4$, Seok Min Kim$^1$, Nam Soo Kim$^1$} 

\address{Department of Electrical and Computer Engineering and INMC, Seoul National University$^1$\\
	Department of Computer Science, Tokyo Institute of Technology$^2$, Google LLC$^3$\\
	Interdisciplinary Program in Cognitive Science, Seoul National University$^4$\\
	\texttt{wicho@hi.snu.ac.kr,sangwhan@iki.fi,prows12@gmail.com}\\\texttt{smkim@hi.snu.ac.kr,nkim@snu.ac.kr}\\}

\abstract{
Paraphrasing is often performed with less concern for controlled style conversion. Especially for questions and commands, style-variant paraphrasing can be crucial in tone and manner, which also matters with industrial applications such as dialog systems. In this paper, we attack this issue with a corpus construction scheme that simultaneously considers the core content and style of directives, namely intent and formality, for the Korean language. Utilizing manually generated natural language queries on six daily topics, we expand the corpus to formal and informal sentences by human rewriting and transferring. We verify the validity and industrial applicability of our approach by checking the adequate classification and inference performance that fit with conventional fine-tuning approaches, at the same time proposing a supervised formality transfer task.
 \\ \newline \Keywords{Paraphrase, Style-variant, Korean, Spoken language, Directives} }

\begin{document}

\maketitleabstract

\section{Introduction}

Paraphrasing, the act of using different sentences with the same meaning \cite{bhagat2013paraphrase}, is strongly related to the text style conversion or transfer \cite{yamshchikov2020style}. While prior studies often modify sentiment or offensiveness \cite{logeswaran2018content,dos2018fighting}, in view of paraphrasing, it should be well checked whether the core content of the sentence is maintained during the conversion process. If the sentence meaning stays the same while changing politeness or formality \cite{rao2018dear}, we can call it paraphrasing or rewriting. Such styles can be represented in diverse ways across genre, domain, and language \cite{jhamtani2017shakespearizing,fu2018style,yang2019generating}. 

Up to date, paraphrasing has been adopted as a useful strategy for text data augmentation. For instance, recent text augmentation schemes such as \newcite{dhole2021nl} exploit various automatic rewriting tools such as abbreviating, transliteration, lexical shift etc., while paraphrasing through text style transfer is one of them. The approach bases on unsupervised learning of English text styles, following the scheme of \newcite{krishna-etal-2020-reformulating}. However, automatic style transfer may not always guarantee the naturalness of the sentence and the preservation of core contents. Also, it is not easy to attain direct text style transfer pairs from unsupervised and automatic approaches. This challenge is visible in the languages with a comparably lower amount of resources, where substantial resources are not guaranteed for each desired text style.

In this light, we attempt to make up a solid scheme for the manual construction of text style transfer database, in a less studied language, Korean. We deal with the scheme of constructing a corpus of style-variant paraphrases for directive sentences such as questions and commands, targeting the Korean language where politeness (suffix) and honorifics play a significant role in conversation \cite{strauss2005indexicality}. Here, we consider topic and speech act as attributes constituting the directive sentence \cite{cho2020discourse} and construct a formal style paraphrase set using the natural language queries displaying each topic and speech act. Finally, style-variant paraphrase pairs are obtained by manual conversion from formal to informal sentences in consideration of content preservation, and are to be released publicly as the first open text style transfer dataset in Korean. Our contribution is as follows: 

\begin{itemize}[noitemsep]
	\item We present a corpus construction scheme capable of performing multiple tasks while enabling parallel sentence style transfer.
	\item We release a Korean corpus where sentence formality style is well defined, regarding the daily used questions and commands.\footnote{\url{https://github.com/cynthia/stylekqc}}
\end{itemize}

\section{Related Work}

In general, sentence style\footnote{In this paper, we view `formality' in Korean as a style, while interchangeably using `conversion' and `transfer'.} is handled regarding tone and manner in writing, though with a subtle difference \cite{brooks2020building}. However, previous research on content-preserving style transfer  \cite{logeswaran2018content,tian2018structured} does not seem to be only about tone in that the change in sentiment may influence the core speaker intent. Furthermore, most approaches were from the perspective of unsupervised learning \cite{dos2018fighting,bao2019generating}, with less explored fields of parallel style-variant corpus for supervised learning, which might provide robust guidance for the generative pre-trained models nowadays \cite{radford2019language}.

This trend was similarly revealed in previous studies on Korean. Since the early approaches follow the studies in English and other languages, sentiment or stance-based style transfer has been predominantly suggested \cite{lee2019controlled,choi2019controlled}.\footnote{Most of the work are not in an internationally readable format; thus, we note here the methods used in the papers.} In \newcite{hong2018korean}, the transfer regarding politeness suffix of the sentence enders was considered at the same time maintaining the sentence meaning, mainly regarding `\textit{hay-yo}' and `\textit{hap-syo}' enders which differ in the degree of formality. However, it dealt only with the syntactic change, not the modification in the lexicon, adverbs, or tone and manner of the speech, which are all considered influential for the honorific system \cite{strauss2005indexicality}. In this regard, we thought that formality style transfer should be well-defined along with content preservation. Furthermore, there is no open dataset for Korean style transfer that can be utilized for research and commercial purposes. We aim to resolve the above issues by proposing a straightforward and effective building scheme.

\section{Proposed Scheme}

We construct a corpus of Korean directives, namely questions and commands, where the question consists of an alternative question (Alt. Q) or wh-question (wh- Q), and the command consists of prohibition (PH) and requirement (REQ), following \newcite{cho2020discourse}. In other words, we target four types of speech acts and assume sentences that can be uttered to humans or artificial intelligent (AI) agents. There are six topics involved in this: \textit{messenger, calendar, weather and news, smart home, shopping,} and \textit{entertainment}, which come from a recent survey on customers' usage \cite{lee2020positioning}. Twelve workers from different backgrounds were recruited. In detail, there were six researchers/students with linguistics background, three researchers/students with non-linguistic background, and three participants working in an industry not related to the linguistics domain. We required specifying two likes and one dislike on the topic, and these preferences were taken into account when creating a total of 6 subgroups with two people each. Here, to help participants interact with each other's strategies and at the same time proceed in a way that is more linguistically feasible, we placed a researcher with the linguistics background to each group. 

We created a construction scheme that goes through the following three steps to check its reliability while generating utterances of 5,000 per topic and 7,500 per speech act.
\begin{enumerate}[noitemsep]
	\item Writing natural language queries
	\item Rewriting queries in a sentence with the formal tone
	\item Converting the formal sentences to informal ones
\end{enumerate}

\paragraph{Query generation} First, query generation is a process in which participants directly suggest the core content of directives which are to be rewritten in a formal style. In this process, participants were asked to write a natural language query for each of the given two speech acts on the assigned topic.\footnote{These were readily provided by the process managers in \newcite{cho2020discourse}, but here we let them be created by the workers to make the contents more diverse and to benefit from the preferences. Also, `query' here does not only apply to the phrase for questions, but also the nominalized phrase for commands. Refer to \newcite{cho-etal-2020-machines} for further information.} Since the query structure differs by speech act type as in \newcite{cho2020discourse}, the created queries did not overlap across the workers. The queries were checked for their suitability, to avoid personally identifiable stuff or those that can cause social harm. 125 queries were generated for each (\textbf{\textit{topic, act}}) pair. The example of queries per some (\textbf{\textit{topic, act}}) is shown below. All the queries are generated in Korean, but described here in English for demonstrative purposes.  
\begin{itemize}[noitemsep]
	\item (\textit{Shopping, Alt. Q}) \textit{The one that has better A/S between Samsung and Apple}
	\item (\textit{Entertainment, Wh-Q}) \textit{The TV channel number where the news is on at 8:00 p.m.}
	\item (\textit{Messenger, PH}) \textit{Not to turn on WeChat automatic update}
	\item (\textit{Smart home, REQ}) \textit{To recharge the wireless vacuum cleaner in the multi-room}
\end{itemize}
No particular principle was considered in the query generation, but the workers were asked to make diverse expressions that fit with colloquial context and daily life. Too knowledge-intensive questions or queries with multiple contents were asked for a modification.

\begin{figure*}
	\centering
	\includegraphics[width=\textwidth]{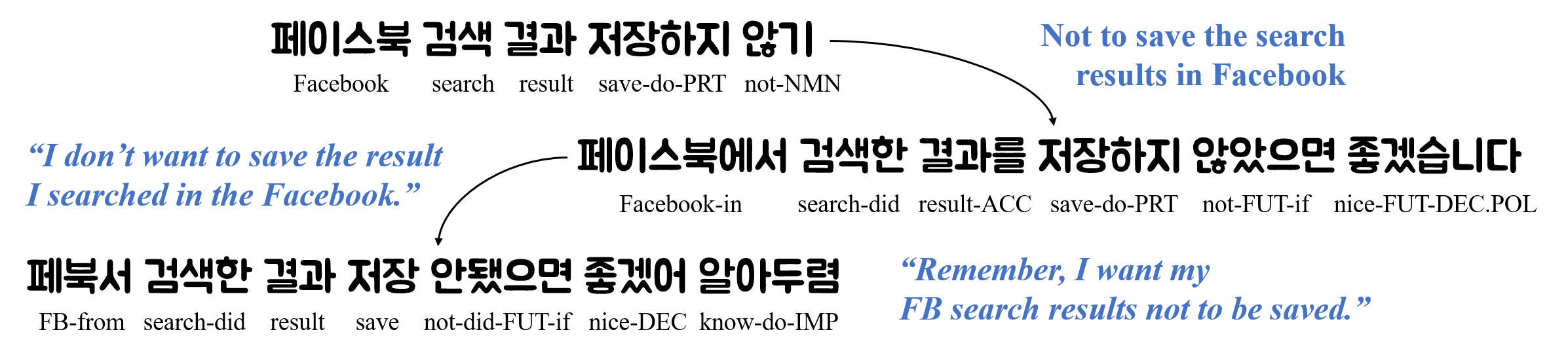}
	\caption{An example of query generation-formal sentence writing-informal transferring, along with the gloss and translation (\textit{PRT} particle, \textit{NMN} nominalizer, \textit{ACC} accusative, \textit{FUT} futuristic, \textit{DEC} declarative, \textit{POL} politeness suffix, \textit{IMP} imperative). Though not reflected in the English translation, the transferring preserves the overall structure of the formal sentence as well as the core content.} \label{fig:fig2}
\end{figure*} 

\paragraph{Writing formal sentences} The next is a process in which the workers of subgroups exchange queries generated by each other and rewrite them into formal style sentences.\footnote{In this process, the workers check the validity of the query created by each other, that the incompleteness of the queries overlooked by the moderator (author) can be pointed out.} We primarily asked for the formal style because there are more diverse expressions for formal utterances in the Korean language regarding indirect speech and honorifics \cite{byon2006role}, so that the paraphrasing is easier compared to informal ones that might not come to the worker's mind at the first place. The formal utterances were required to fit with the conversation with senior or elderly addressees rather than friends or juniors. 

Rewriting was required for a total of 5 sentences. To make the paraphrases as diverse as possible, the asking strategies in \newcite{byon2006role} and \newcite{cho2008strategic} were requested. We display some excerpts:
\begin{itemize}[noitemsep]
	\item Softening the commands to requests
	\item Indirectly mentioning the addressee's obligation
	\item Mentioning the addressee's responsibility
	\item Alleviating the addressee's burden with polarity items such as \textit{please} or \textit{bit}
	\item Asking the availability of the addressee
\end{itemize}
Some of these characteristics are shared across the culture \cite{brown1987politeness}. It may also be exhibited similar in the East Asian society \cite{gu1990politeness} and within a similar syntax such as Japanese \cite{okamoto1999situated,fukada2004universal}. However, we faced language-specific considerations regarding functional and lexical expressions and asked the workers to reflect them in the construction. Simultaneously, to fit with the naturalness within colloquial context, written-style or outdated phrases/words were avoided.

\paragraph{Converting to informal style} The final process is modifying directive sentences written in formal style into informal sentences. Here, the workers convert the other person's formal sentences, created from the original query they had generated, checking the typos and misunderstandings once again. `Informality' defined here is slightly different from being rude or impolite, but instead means that  the conversation moves towards a more comfortable and personal relationship. \cite{rao2018dear}. 

In this process, we asked the workers to maintain the overall sentence structure, of which the diversity was already obtained owing to policies in writing formal sentences. With this, we could prevent the potential overlap between the converted sentences and also guarantee the `parallelness' of the created data. This can be more effective in the Korean language where indirectness is often distinguished from formality; for instance, a cautious request to a younger brother can be informal but indirect. 

Style conversion was performed in various aspects such as change in sentence enders, honorifics, and lexicons (such as \textit{nation} to \textit{country}). The workers were encouraged to insert or delete some phrases depending on the naturalness of the content, and to perform at least two word-level modifications. The detailed guideline\footnote{ \url{https://docs.google.com/document/d/1gjyEMCcp0mxmdzSKdd5OrLFVikyq22OsxXHisSr2THY} written in Korean.} for the whole process was provided to the workers with example query-sentence tuples, and we exhibit one of them (Figure 1).


\paragraph{Refinement} The corpus was refined by three native speakers with corpus construction experience for Korean directive sentences.
In this process, typos, awkward sentences, and paraphrases that are not sufficiently diverse were inspected, and the reviews were reflected by the moderator.


\section{Experiment}

\subsection{Task Setting}

Through the experiment, we display that the proposed construction scheme provides a corpus that enables creating multiple task sets simultaneously, which can bring advantages from a practical viewpoint.
\begin{itemize}[noitemsep]
	\item Topic classification
	\item Speech act classification
	\item Paraphrase detection
	\item Sentence style transfer
\end{itemize}

\subsection{Implementation}

For each of the total 24 [topic, act] chunks where we have 125 queries each, we set aside 80\% (100 queries) for training, 4\% (5 queries) for validation, and 16\% (20 queries) for the test. From the whole dataset of volume 30,000, the training set contains 24,000 sentences and 1,200/4,800 for dev/test each. The queries were chosen randomly, and all the sets have an equal rate of topic and speech act ratio.  

Topic (TOPIC) and speech act (ACT) classification are intuitively formulated. There are 5,000 utterances for each topic and 7,500 utterances for each speech act, where six topics and four speech act types are set as labels. 

Paraphrase detection (PARA) requires a sentence pair. In \newcite{cho2020discourse}, the sentence similarity was defined  5-fold, checking if the topic or speech act overlaps between the two input sentences, with the highest similarity if the queries are identical (the paraphrases). The paraphrase detection task was derived by formulating the multi-class problem into a binary task. See \texttt{DATA-GEN/mkdata.py} in the supplementary material\footnote{\url{https://www.dropbox.com/s/ju53oan78u2nfkx/supple-data.zip?dl=0}} for further detail. 

Finally, we checked whether sentence style transfer (STYLE) works using the pairs within; 12,000 pairs for training, 600 for validation and 2,400 for the test. The training was done in the way of converting the formal sentences to informal ones. 

Both sentence classification and paraphrase detection tasks were implemented based on a BERT-based \cite{devlin2019bert} KcBERT\footnote{\url{https://github.com/Beomi/KcBERT}} \cite{lee2020kcbert}, and for sentence style transfer, KoGPT2\footnote{\url{https://github.com/SKT-AI/KoGPT2}} that bases on GPT2 \cite{radford2019language} was adopted. F1 (macro) and accuracy were used for the classification tasks, and for style transfer, we checked character edit distance (CED). The accuracy for style transfer ($\dagger$) denotes the precision obtained with the model learned upon the train set \cite{pang2019daunting}. Experimental settings are provided as supplementary.

\begin{table}[]
	\centering
	\resizebox{\columnwidth}{!}{%
		\begin{tabular}{ccccc}
			\hline
			& \textbf{TOPIC} & \textbf{ACT} & \textbf{PARA} & \textbf{STYLE} \\ \hline
			\textbf{Input}    & Sentence       & Sentence     & Pair          & Sentence       \\
			\textbf{Class \#} & 6              & 4            & 2             & -              \\
			\textbf{Volume}   & 30,000         & 30,000       & 270,000       & 15,000         \\
			\textbf{F1 Score} & 92.68          & 97.75        & 99.93         & -              \\
			\textbf{Accuracy} & 92.83          & 97.75        & 99.93         & 99.58$\dagger$              \\
			\textbf{CED}  & -              & -            & -             & 0.451    \\ \hline
		\end{tabular}%
	}
	\caption{Experiment results on four subtasks.}
	\label{tab:my-table}
\end{table}

\subsection{Results}

In classification and inference, we have the evaluation results that show consistency between the train and test dataset (Table 1). Considering that queries in each set are distinguished from each other, we claim that our dataset displays the extensibility to wider world problems, also providing the comprehensive coverage of topics and acts that are of interest in usual conversation and smart speaker dialogues. Though the baseline score is quite high for ACT and PARA, it does not harm one of our goals to provide a solid scheme for corpus construction that suffices practical, real-world applicability.

On STYLE, we adopted CED since our `style' more regards the change in suffix and some lexicons rather than the whole word order and phrase usage.\footnote{On using other objective measures, the morpheme-level tokenization is not yet unified for Korean sentences, to make evaluation harder. More explanation is available in Appendix~\ref{app:ced}.} Nonetheless, we found the transfer task still challenging in view of the objective measure. Instead, we observed the practical validity using a style classifier learned upon train and valid set, which displays sufficiently high accuracy.\footnote{More explanation on using classifier accuracy in style checking is available in Appendix~\ref{app:acc}.} We qualitatively checked that the seq2seq \cite{sutskever2014sequence} approach with a pre-trained generative model guarantees the intended style transfer.

\subsubsection{Error Analysis}
For style transfer, some errors have occurred in the following forms:

\begin{enumerate}
    \item Unknown stop in the decoding session
    \item Repetition of some phrases
    \item Appearance of irrelevant terms
\end{enumerate}

\paragraph{Unknown stops}
We first assumed OOV for a reason, but it turned out not since it happened for the text cases where all the tokens exist in the training set. Another analysis suggests that the change of word order (which is tolerated in Korean for being scrambling) which makes it challenging for the language understanding module to comprehend a full sentence, might have caused the decoding module to fall in collapse and finish the decoding just by facing the end of usual sentences. For instance, an in-out pair\medskip\\(a) 국내 브랜드가 더 많이 들어가 있는 곳을 알아봐주세요 지마켓과 신세계 중에 (``Please find out where more domestic brands are located,  among G-Market and Shinsegae.")\\(b) *\footnote{Wrong sentence.}국내 브랜드가 더 많이 들어가 있는 곳 좀 알아봐줘 지마켓 (*``G-Market, find out where more domestic brands are in.")\medskip\\ shows that the scrambling, which preserves sentence acceptability in Korean, might confuse the trained module.

\paragraph{Repitition}
The repetition of phrases bursts out when the model is confused about what to transfer, sometimes because it misunderstood the act of the utterance. For instance, in an in-out pair\medskip\\ (c) 내일 재고확인하세요 모레 재고확인하세요 ("Will you check stock tomorrow or the day after tomorrow?")	\\(d) *내일 재고확인 좀 해 모레 재고확인해야겠어 모레 재고확인해야겠어 (*"Check stock tomorrow. I will check it the day after tomorrow. I will check it the day after tomorrow.")\medskip\\ the transfer model fails to understand that the input sentence is an alternative question and transfers it as a command (due to a seemingly ambiguous sentence ender 요 (yo) - whose role is clear at this circumstance), finally displaying a repetition, failing to emit an acceptable sentence.

\paragraph{Irrelevant terms}
The appearance of irrelevant terms happened rarely, but mainly seemed to be owing to the knowledge within the generative pre-trained models. It would be our future work to lessen this kind of malfunction where the pre-trained bias negatively affects the fine-tuned model.


\subsubsection{Discussion}

We have some notes on the validity of the created dataset. Primarily, though the dataset is first suggested open corpus for Korean style transfer, the granularity of the style difference within the pair is not provided here as in \newcite{rao2018dear}. Also, since our dataset provides the style transfer that maintains the overall sentence structure, some sentence pairs show minor differences, which is sufficient for spoken language processing but less robust to digitized online texts. Finally, since the formality conversion regards morpho-syntactic and lexical changes rather than the paraphrasing done in writing the formal sentences, the style diversity of expressions is limited to the sentence formats that are not awkward to utter.

Despite the limitations, we want to emphasize that our approach can suggest a reliable and efficient scheme for the service providers or task managers aiming at a particular style transfer for various types of sentences. For instance, if one replaces input queries with some structured query language (SQL) or canonical forms of statements and use `rudeness' or `twitter-likeness' as a style, the parallel dataset can be created in the same way, with a slightly different guideline. This kind of pair generation has been done with rule or back translation in \newcite{rao2018dear}, but we believe that human-aided construction is more reliable and eventually reduces the necessity of additional human checking. Also, see Appendix~\ref{app:ethics} to see how our manual construction process has considered the ethical sides of human factors.

\section{Conclusion}

In this paper, we construct and disclose the first style-variant Korean paraphrase corpus. Topic, speech act, and paraphrase are simultaneously considered in evaluating the final corpus, where the consistent composition is assumed to be guaranteed by the evaluation results. The entire guideline is currently specific to the formality transfer in Korean, but can be utilized in making up other parallel style transfer corpus with an extended pool of topics, speech acts, queries, and style. All the resources are available online\footnote{\url{https://github.com/cynthia/stylekqc}}, and we provide another implementation for politeness transfer using a Korean public PLM\footnote{\url{https://colab.research.google.com/drive/1YjU1wlwl26X49hQLr6ZQvOKQm2yiR4sg}} to facilitate the future research on Korean text style transfer.

\section{Acknowledgements}

This work was supported by Institute of Information \& communications Technology Planning \& Evaluation (IITP) grant funded by the Korea government(MSIT) (No.2021-0-00456, Development of Ultra-high Speech Quality  Technology for Remote Multi-speaker Conference System). We appreciate Seonghyun Kim for providing a codebase for KoBART implementation. Also, the corpus construction was possible thanks to the help of twelve passionate participants, namely Kyung Seo Ki, Dongho Lee, Yoon Kyung Lee, Hee Young Park, Yulhee Kim, Seyoung Park, Jiwon An, Jeonghwa Cho, Kihyo Park, Kyuhwan Lee, Soomin Lee, and Minhwa Chung.











\section{Bibliographical References}\label{reference}

\bibliographystyle{lrec2022-bib}
\bibliography{lrec2022-example,anthology,custom}

\begin{thebibliography}{}

\bibitem[\protect\citename{Bao \bgroup et al.\egroup }2019]{bao2019generating}
Bao, Y., Zhou, H., Huang, S., Li, L., Mou, L., Vechtomova, O., Dai, X., and
  Chen, J.
\newblock (2019).
\newblock Generating sentences from disentangled syntactic and semantic spaces.
\newblock In {\em Proceedings of the 57th Annual Meeting of the Association for
  Computational Linguistics}, pages 6008--6019.

\bibitem[\protect\citename{Bhagat and Hovy}2013]{bhagat2013paraphrase}
Bhagat, R. and Hovy, E.
\newblock (2013).
\newblock What is a paraphrase?
\newblock {\em Computational Linguistics}, 39(3):463--472.

\bibitem[\protect\citename{Brooks}2020]{brooks2020building}
Brooks, C.
\newblock (2020).
\newblock {\em Building Blocks of Academic Writing}.
\newblock BCcampus.

\bibitem[\protect\citename{Brown \bgroup et al.\egroup
  }1987]{brown1987politeness}
Brown, P., Levinson, S.~C., and Levinson, S.~C.
\newblock (1987).
\newblock {\em Politeness: Some universals in language usage}, volume~4.
\newblock Cambridge University Press.

\bibitem[\protect\citename{Byon}2006]{byon2006role}
Byon, A.~S.
\newblock (2006).
\newblock The role of linguistic indirectness and honorifics in achieving
  linguistic politeness in {K}orean requests.
\newblock {\em Journal of Politeness Research}, 2(2):247--276.

\bibitem[\protect\citename{Cho \bgroup et al.\egroup }2020a]{cho2020discourse}
Cho, W.~I., Kim, J.~I., Moon, Y.~K., and Kim, N.~S.
\newblock (2020a).
\newblock Discourse component to sentence ({DC2S}): An efficient human-aided
  construction of paraphrase and sentence similarity dataset.
\newblock In {\em Proceedings of The 12th Language Resources and Evaluation
  Conference}, pages 6819--6826.

\bibitem[\protect\citename{Cho \bgroup et al.\egroup
  }2020b]{cho-etal-2020-machines}
Cho, W.~I., Moon, Y., Moon, S., Kim, S.~M., and Kim, N.~S.
\newblock (2020b).
\newblock Machines getting with the program: Understanding intent arguments of
  non-canonical directives.
\newblock In {\em Findings of the Association for Computational Linguistics:
  EMNLP 2020}, pages 329--339, Online, November. Association for Computational
  Linguistics.

\bibitem[\protect\citename{Cho}2008]{cho2008strategic}
Cho, Y.
\newblock (2008).
\newblock Strategic use of {K}orean honorifics functions of
  ‘partner-deference sangdae-nopim’.
\newblock {\em Dialogue and Rhetoric}, 2:155.

\bibitem[\protect\citename{Choi and Na}2019]{choi2019controlled}
Choi, H.-J. and Na, S.-H.
\newblock (2019).
\newblock Delete and generate: {K}orean style transfer based on deleting and
  generating word n-grams.
\newblock In {\em Annual Conference on Human and Language Technology}, pages
  400--403. Human and Language Technology.

\bibitem[\protect\citename{Devlin \bgroup et al.\egroup }2019]{devlin2019bert}
Devlin, J., Chang, M.-W., Lee, K., and Toutanova, K.
\newblock (2019).
\newblock {BERT}: Pre-training of deep bidirectional transformers for language
  understanding.
\newblock In {\em Proceedings of the 2019 Conference of the North American
  Chapter of the Association for Computational Linguistics: Human Language
  Technologies, Volume 1 (Long and Short Papers)}, pages 4171--4186.

\bibitem[\protect\citename{Dhole \bgroup et al.\egroup }2021]{dhole2021nl}
Dhole, K.~D., Gangal, V., Gehrmann, S., Gupta, A., Li, Z., Mahamood, S.,
  Mahendiran, A., Mille, S., Srivastava, A., Tan, S., Wu, T., Sohl-Dickstein,
  J., Choi, J.~D., Hovy, E.~H., Dusek, O., Ruder, S., Anand, S., Aneja, N.,
  Banjade, R., Barthe, L., Behnke, H., Berlot-Attwell, I., Boyle, C., Brun,
  C.~D., Cabezudo, M. A.~S., Cahyawijaya, S., Chapuis, E., Che, W., Choudhary,
  M., Clauss, C., Colombo, P., Cornell, F., Dagan, G., Das, M., Dixit, T.,
  Dopierre, T., Dray, P.-A., Dubey, S., Ekeinhor, T., Giovanni, M.~D., Gupta,
  R., Hamla, L., Han, S., Harel-Canada, F., Honor{\'e}, A., Jindal, I., Joniak,
  P.~K., Kleyko, D., Kovatchev, V., Krishna, K., Kumar, A., Langer, S., Lee,
  S.~R., Levinson, C.~J., Liang, H., Liang, K., Liu, Z., Lukyanenko, A.,
  Marivate, V., de~Melo, G., Meoni, S., Meyer, M., Mir, A., Moosavi, N.~S.,
  Muennighoff, N., Mun, T. S.~H., Murray, K.~W., Namysl, M., Obedkova, M., Oli,
  P., Pasricha, N., Pfister, J., Plant, R., Prabhu, V.~U., Pais, V.~F., Qin,
  L., Raji, S., Rajpoot, P.~K., Raunak, V., Rinberg, R., Roberts, N.,
  Rodriguez, J.~D., Roux, C., VasconcellosP.H., S., Sai, A.~B., Schmidt, R.~M.,
  Scialom, T., Sefara, T.~J., Shamsi, S., Shen, X., Shi, H., Shi, Y., Shvets,
  A.~V., Siegel, N., Sileo, D., Simon, J., Singh, C., Sitelew, R., Soni, P.,
  Sorensen, T.~M., Soto, W., Srivastava, A., Srivatsa, K. V.~A., Sun, T.,
  MukundVarma, T., Tabassum, A., Tan, F.~A., Teehan, R., Tiwari, M., Tolkiehn,
  M., Wang, A., Wang, Z., Wang, G., Wang, Z.~J., Wei, F., Wilie, B., Winata,
  G.~I., Wu, X., Wydma'nski, W., Xie, T., Yaseen, U., Yee, M.-H., Zhang, J.,
  and Zhang, Y.
\newblock (2021).
\newblock Nl-augmenter: A framework for task-sensitive natural language
  augmentation.
\newblock {\em arXiv preprint arXiv:2112.02721}.

\bibitem[\protect\citename{dos Santos \bgroup et al.\egroup
  }2018]{dos2018fighting}
dos Santos, C., Melnyk, I., and Padhi, I.
\newblock (2018).
\newblock Fighting offensive language on social media with unsupervised text
  style transfer.
\newblock In {\em Proceedings of the 56th Annual Meeting of the Association for
  Computational Linguistics (Volume 2: Short Papers)}, pages 189--194.

\bibitem[\protect\citename{Fu \bgroup et al.\egroup }2018]{fu2018style}
Fu, Z., Tan, X., Peng, N., Zhao, D., and Yan, R.
\newblock (2018).
\newblock Style transfer in text: Exploration and evaluation.
\newblock In {\em AAAI}, pages 663--670.

\bibitem[\protect\citename{Fukada and Asato}2004]{fukada2004universal}
Fukada, A. and Asato, N.
\newblock (2004).
\newblock Universal politeness theory: application to the use of {J}apanese
  honorifics.
\newblock {\em Journal of Pragmatics}, 36(11):1991--2002.

\bibitem[\protect\citename{Gu}1990]{gu1990politeness}
Gu, Y.
\newblock (1990).
\newblock Politeness phenomena in modern {C}hinese.
\newblock {\em Journal of Pragmatics}, 14(2):237--257.

\bibitem[\protect\citename{Hong \bgroup et al.\egroup }2018]{hong2018korean}
Hong, T., Xu, G., Ahn, H., Kang, S., and Seo, J.
\newblock (2018).
\newblock Korean text style transfer using attention-based sequence-to-sequence
  model.
\newblock In {\em Annual Conference on Human and Language Technology}, pages
  567--569. Human and Language Technology.

\bibitem[\protect\citename{Jhamtani \bgroup et al.\egroup
  }2017]{jhamtani2017shakespearizing}
Jhamtani, H., Gangal, V., Hovy, E., and Nyberg, E.
\newblock (2017).
\newblock Shakespearizing modern language using copy-enriched sequence to
  sequence models.
\newblock In {\em Proceedings of the Workshop on Stylistic Variation}, pages
  10--19.

\bibitem[\protect\citename{Krishna \bgroup et al.\egroup
  }2020]{krishna-etal-2020-reformulating}
Krishna, K., Wieting, J., and Iyyer, M.
\newblock (2020).
\newblock Reformulating unsupervised style transfer as paraphrase generation.
\newblock In {\em Proceedings of the 2020 Conference on Empirical Methods in
  Natural Language Processing (EMNLP)}, pages 737--762, Online, November.
  Association for Computational Linguistics.

\bibitem[\protect\citename{Lee \bgroup et al.\egroup }2019]{lee2019controlled}
Lee, J., Oh, Y., Byun, h., and Min, K.
\newblock (2019).
\newblock Controlled {K}orean style transfer using {BERT}.
\newblock In {\em Annual Conference on Human and Language Technology}, pages
  395--399. Human and Language Technology.

\bibitem[\protect\citename{Lee \bgroup et al.\egroup }2020]{lee2020positioning}
Lee, J.~H., Seon, H.~J., and Lee, H.~J.
\newblock (2020).
\newblock Positioning of smart speakers by applying text mining to consumer
  reviews: Focusing on artificial intelligence factors.
\newblock {\em Knowledge Management Research}, 21(1):197--210.

\bibitem[\protect\citename{Lee}2020]{lee2020kcbert}
Lee, J.
\newblock (2020).
\newblock Kcbert: Korean comments bert.
\newblock In {\em Annual Conference on Human and Language Technology}. Human
  and Language Technology.

\bibitem[\protect\citename{Logeswaran \bgroup et al.\egroup
  }2018]{logeswaran2018content}
Logeswaran, L., Lee, H., and Bengio, S.
\newblock (2018).
\newblock Content preserving text generation with attribute controls.
\newblock In {\em Advances in Neural Information Processing Systems}, pages
  5103--5113.

\bibitem[\protect\citename{Okamoto}1999]{okamoto1999situated}
Okamoto, S.
\newblock (1999).
\newblock Situated politeness: Manipulating honorific and non-honorific
  expressions in {J}apanese conversations.
\newblock {\em Pragmatics}, 9(1):51--74.

\bibitem[\protect\citename{Pang}2019]{pang2019daunting}
Pang, R.~Y.
\newblock (2019).
\newblock The daunting task of real-world textual style transfer
  auto-evaluation.
\newblock {\em arXiv preprint arXiv:1910.03747}.

\bibitem[\protect\citename{Radford \bgroup et al.\egroup
  }2019]{radford2019language}
Radford, A., Wu, J., Child, R., Luan, D., Amodei, D., and Sutskever, I.
\newblock (2019).
\newblock Language models are unsupervised multitask learners.
\newblock {\em OpenAI blog}, 1(8):9.

\bibitem[\protect\citename{Rao and Tetreault}2018]{rao2018dear}
Rao, S. and Tetreault, J.
\newblock (2018).
\newblock Dear sir or madam, may i introduce the {GYAFC} dataset: Corpus,
  benchmarks and metrics for formality style transfer.
\newblock In {\em Proceedings of the 2018 Conference of the North American
  Chapter of the Association for Computational Linguistics: Human Language
  Technologies, Volume 1 (Long Papers)}, pages 129--140.

\bibitem[\protect\citename{Strauss and Eun}2005]{strauss2005indexicality}
Strauss, S. and Eun, J.~O.
\newblock (2005).
\newblock Indexicality and honorific speech level choice in {K}orean.
\newblock {\em Linguistics}, 43(3):611--651.

\bibitem[\protect\citename{Sutskever \bgroup et al.\egroup
  }2014]{sutskever2014sequence}
Sutskever, I., Vinyals, O., and Le, Q.~V.
\newblock (2014).
\newblock Sequence to sequence learning with neural networks.
\newblock In {\em Advances in Neural Information Processing Systems}, pages
  3104--3112.

\bibitem[\protect\citename{Tian \bgroup et al.\egroup
  }2018]{tian2018structured}
Tian, Y., Hu, Z., and Yu, Z.
\newblock (2018).
\newblock Structured content preservation for unsupervised text style transfer.
\newblock {\em arXiv preprint arXiv:1810.06526}.

\bibitem[\protect\citename{Yamshchikov \bgroup et al.\egroup
  }2020]{yamshchikov2020style}
Yamshchikov, I., Shibaev, V., Khlebnikov, N., and Tikhonov, A.
\newblock (2020).
\newblock Style-transfer and paraphrase: Looking for a sensible semantic
  similarity metric.
\newblock {\em arXiv preprint arXiv:2004.05001}.

\bibitem[\protect\citename{Yang \bgroup et al.\egroup
  }2019]{yang2019generating}
Yang, Z., Cai, P., Feng, Y., Li, F., Feng, W., Chiu, E.-Y., and Yu, H.
\newblock (2019).
\newblock Generating classical chinese poems from vernacular {C}hinese.
\newblock In {\em Proceedings of the Conference on Empirical Methods in Natural
  Language Processing. Conference on Empirical Methods in Natural Language
  Processing}, volume 2019, page 6155. NIH Public Access.

\end{thebibliography}

\medskip

\appendix

\section{Analysis on STYLE Results}
\label{app:style}

In STYLE, CED and accuracy are used, where each is defined a little bit different from usual cases.

\subsection{Using CED}\label{app:ced} First, on CED, since the output length may differ from GT, we normalize the CED with length so that it comes between 0 and 1, such as defined in \texttt{SOURCE/EVAL-STYLE/eval\_style.py} among the supplementary material. In Korean, \textit{character} denotes a morpho-syllabic block that corresponds to the subword in English, thus CED can have a role as a subword-level edit distance. This was considered more appropriate than BLEU or METEOR, which are usual for other Latin alphabet-based style transfer studies, since 1) ours aims at a structure-conservative style-variant paraphrasing, and 2) morphological decomposition schemes are not solidly unified in the empirical studies. Besides, we did not choose semantic-level measures such as BERTScore since most of the outputs would record a high score because the paraphrasing was guaranteed. 

\subsection{Using Accuracy}\label{app:acc} On the accuracy, which is defined differently from TOPIC, ACT, or PARA since the aim of STYLE is not originally in making a classifier, we check if the style classifier trained with the samples of the training set can precisely classify the transferred test sentences as informal ones. Thus, only the accuracy, which equals precision in this scenario, is calculated. Achieving a high performance here indirectly shows that the style transfer is adequately performed for a large portion of scenarios.

\section{Ethical Considerations}
\label{app:ethics}

In the corpus construction procedure which bases upon the documented approval of the workers, adequate compensation was paid to each of them, in all the processes of query generation, writing formal sentences, and transferring them to the informal one. 

The participants, recruited from social media and the web, are familiar with smart speakers, and some of them had experience in corpus construction processes. For 12 participants, 250 WON ($\approx$\$0.22) was provided in writing each query and 200 WON ($\approx$\$0.18) for making up the sentences. Thus, each participant was paid 600,000 WON ($\approx$\$540) to make up 250 queries and write 2,500 sentences. 

Our resource is free from license issues since all the materials were created according to the guideline (a kind of template) and checked for post-processing. The outcome of our project does not contain any personally identifiable information, nor the contents that can induce social harm.

\end{document}